\documentclass[wcp]{jmlr}

\usepackage{longtable}

\usepackage{booktabs}

\makeatletter
\let\Ginclude@graphics\@org@Ginclude@graphics 
\makeatother

\jmlrvolume{222}
\jmlryear{2023}
\jmlrworkshop{ACML 2023}

\usepackage{bbm}
\usepackage{amsfonts}
\usepackage{times}
\usepackage{hyperref}

\usepackage{graphicx}
\usepackage{xspace}
\usepackage{multirow}

\usepackage{paralist}
\usepackage{mismath}

\usepackage[utf8]{inputenc} 
\usepackage[T1]{fontenc}    
\usepackage{wrapfig,lipsum}
\usepackage{nicefrac}       
\usepackage{microtype}      

\newcommand{\whatb}[1]{\textcolor{blue}{#1}}
\newenvironment{myfont}{\fontfamily{pcr}\selectfont}{\par}
\DeclareTextFontCommand{\textmyfont}{\myfont}

\definecolor{darker}{rgb}{0.357,0.608,0.835}
\definecolor{dark}{rgb}{0.824,0.87,0.937}
\definecolor{light}{rgb}{0.917,0.937,0.968}
\definecolor{subtotal}{rgb}{0.76, 0.8, 0.88}
\definecolor{total}{rgb}{0.6, 0.75, 0.88}
\SetKwComment{Comment}{// }{}
\RestyleAlgo{ruled}

\title[Temporal Event set Modeling]{Deep Representation Learning for Prediction of Temporal Event Sets\\in the Continuous Time Domain}

\author{\Name{Parag Dutta} \Email{paragdutta@iisc.ac.in}\\
  \Name{Kawin Mayilvaghanan} \Email{kawinm@iisc.ac.in}\\
  \Name{Pratyaksha Sinha} \Email{pratyaksha1@iisc.ac.in}\\
  \Name{Ambedkar Dukkipati} \Email{ambedkar@iisc.ac.in}\\
  \addr Department of Computer Science and Automation\\
  Indian Institute of Science (IISc), Bangalore, KA, IN - 560012
}

\editors{Berrin Yan{\i}ko\u{g}lu and Wray Buntine}

\begin{document}

\maketitle

\begin{abstract}
Temporal Point Processes (TPP) play an important role in predicting or forecasting events. Although these problems have been studied extensively, predicting multiple simultaneously occurring events can be challenging. For instance, more often than not, a patient gets admitted to a hospital with multiple conditions at a time. Similarly people buy more than one stock and multiple news breaks out at the same time. Moreover, these events do not occur at discrete time intervals, and forecasting event sets in the continuous time domain remains an open problem. 
Na\"ive approaches for extending the existing TPP models for solving this problem lead to dealing with an exponentially large number of events or ignoring set dependencies among events. In this work, we propose a scalable and efficient approach based on TPPs to solve this problem. Our proposed approach incorporates contextual event embeddings, temporal information, and domain features to model the temporal event sets. We demonstrate the effectiveness of our approach through extensive experiments on multiple datasets, showing that our model outperforms existing methods in terms of prediction metrics and computational efficiency.
To the best of our knowledge, this is the first work that solves the problem of predicting event set intensities in the continuous time domain by using TPPs.\footnote[1]{In proceedings of ACML 2023}
\end{abstract}
\begin{keywords}
Temporal Point Processes, Self-supervised learning, Forecasting, Events
\end{keywords}


\section{Introduction}
In today's complex and dynamic world, the need for accurate and reliable predictions is greater than ever. By making event predictions, we can identify potential risks and opportunities and take appropriate action to prepare for or capitalize on them. Event prediction problems have been studied in machine learning literature extensively, where the approaches range from sequence modeling to temporal point processes. Almost every approach deals with the problem of predicting a single event based on historical data. On the other hand, many practical problems require forecasting multiple events (a set of events), which inevitably requires us to model the distribution of such event sets over time because a simple prediction of whether an event set will occur is insufficient. In medical diagnosis, for instance, it is critical to know whether a particular condition is present and when it is likely to occur next so that preventive measures are taken accordingly (refer to Figure \ref{Figure: Typical Dataset Example}). Another example is trying to predict when and what set of items will a person check-out on an e-commerce website. Prior knowledge of the same can help reduce the shipment charges if the items are placed at convenient locations beforehand. Solution to event set prediction problem can provide valuable insights into the underlying patterns in the data and can be useful for identifying trends and making long-term predictions about the future.

Although multi-variate temporal event modeling has been explored before, these methods are rendered ineffective for modeling temporal event sets \citep{mvhawkes, nhp, thawkes}. One may try to modify existing approaches for predicting temporal event sets. Considering all combinations of events as unique events can be one way to model the problem and still use the existing temporal event modeling approaches. However, the number of events increases exponentially and is impractical. An alternate approach is to decompose the event set into multiple singleton elements, assign the event set timestamp to each event of the event set individually, and then model them as regular temporal events. This approach, although tractable, does not consider the relations and dependencies among the events in the event sets.

In this paper, we propose a new approach based on deep representation learning that can resolve all the above-mentioned problems. Our \textit{contributions} are as follows:

\begin{figure}[t]
    \centering
    \includegraphics[width=\linewidth]{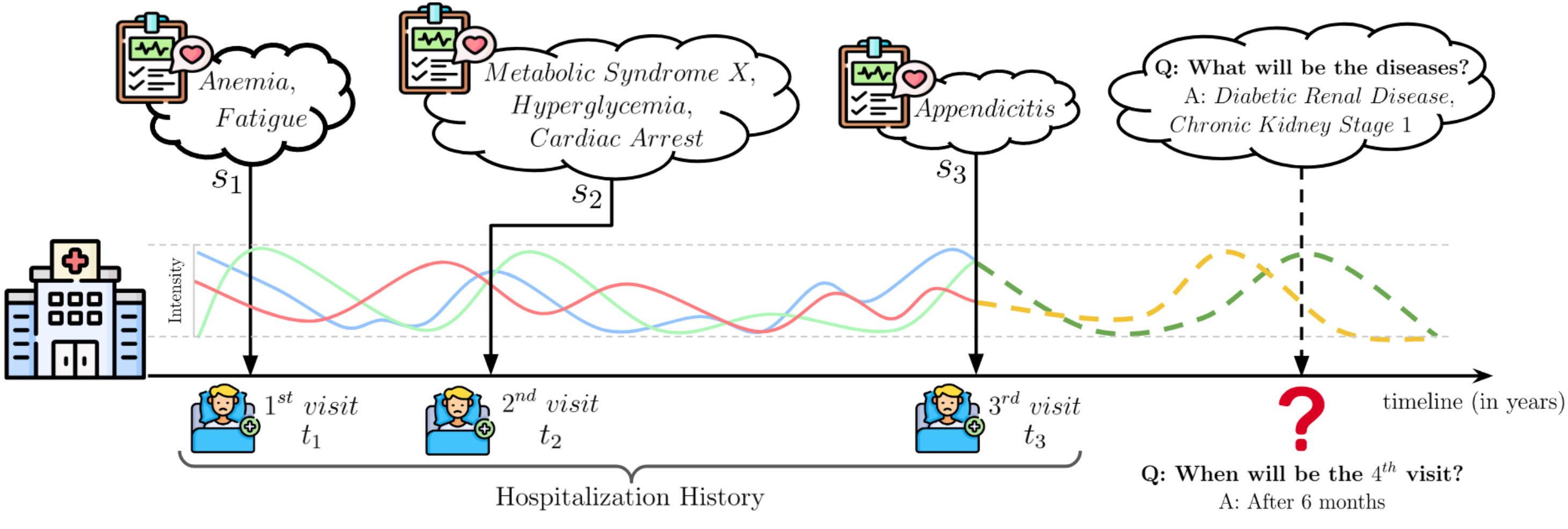}
    \label{Figure: Typical Dataset Example}
    \caption{\small A typical temporal event set data sequence $\mathcal{S}$. Temporal Event set Modeling aims to predict both the event sets and the time of its occurrence given the corresponding history in the continuous time domain. For instance, as shown in the figure, given hospitalization history, we predict when and with what diseases/conditions the patient might be hospitalized in the future.}
\end{figure}


\begin{compactitem}
    \item [1.] We propose a Contextual Self-Supervised Contrastive Learning objective for training an \textit{Event-Encoder}, which learns representations of events in event sets.
    \item [2.] We propose TESET, a Temporal Event set modeling framework that uses event set embeddings and combines them sequentially using transformer-based models.
    \item [3.] We utilize intensity and temporal prediction heads to predict the intensity distribution of the event set along with the time of occurrence.
    \item [4.] In our approach, we also facilitate using domain-specific features for learning better representations.
\end{compactitem}




\section{Related Works}
\label{Section:RelatedWorks}

Classic temporal event modeling works include Gaussian Processes \citep{gaussian_proc} and Multi-variate Hawkes Processes \citep{mvhawkes}, as mentioned earlier. To deal with the parametric kernels of the Hawkes process, \cite{nhp} proposed the Neural Hawkes Process, which can use the expressive power of LSTMs to learn the intensities. Transformer Hawkes Process \citep{thawkes} is another work that tries to use the computational efficiency of Transformers and the self-attention mechanism to solve RNNs' inability to learn long-term dependencies.



While the work of~\cite{chawkes} tries to model the patient's EHR, it does not consider the patient having multiple codes in the same visit. For the same task, \citep{tgram} uses a graph-based model to show improvements over simple RNN-based methods~\citep{dr}.
The work of~\cite{gbert} uses a variant of the Masked Language Modelling (MLM) objective as a pre-training task. Similarly, for recommendations, \citep{sasr} uses attention-based models, and \citep{b4r} uses an MLM.


BEHRT~\citep{behrt} depicts diagnoses as words, visits as sentences, and a patient’s entire medical history as a document to use multi-head self-attention, positional encoding, and MLM for EHR. Med-Bert~\citep{medbert} further adds to this concept by using serialization embeddings (order of codes within each visit using prior knowledge) besides code embeddings and positional (visit) encodings.
Bert4Rec~\citep{b4r} uses an MLM-like bidirectional pre-training for predicting user-item interactions. Transformer4Rec~\citep{t4r} further uses session information to enhance the previous works.

Recent works in set modeling as Sets2Sets \citep{sets2sets} propose an encoder-decoder framework to predict event sets at discrete time steps, where the event set representation is obtained by aggregating the corresponding event embeddings by average pooling. DSNTSP \citep{dsntsp} uses a transformer framework to learn item and set representations and captures temporal dependencies separately.

However, it must be noted that all the aforementioned methods either lack the ability to encode sets or they are applicable only for a discrete-time setting. To the best of our knowledge, the work proposed in this paper is the first to models event sets in the continuous time domain and solve the forecasting problem using TTPs.



\section{Proposed Approach}
\label{Section: Proposed Approaches}
We propose a two-step representation learning approach in Section~\ref{SubSection: Self-supervised Pre-training} and \ref{SubSection: Hierarchical training} for modeling the temporal event sets. The pre-trained representation model thus obtained after the two steps of training can then be fine-tuned for the required downstream tasks.

\subsection{Notations and Preliminaries}

Let $\mathcal{S}$ denote an input sequence. Each element $\mathbf{s}_k \in \mathcal{S}$ corresponds to an event set and is ordered chronologically, where $k \in [|\mathcal{S}|]$ ($|\mathbf{x}|$ counts the number of elements in the set $\mathbf{x}$ and $[n]$ represents the set $\{1,2,...,n\}$). $\mathbf{s}_k \subset \mathcal{I}$ is a set of events, where $\mathcal{I}$ is the set of all possible events. Every $\mathbf{s}_k$ has an associated timestamp and (optionally) a set of features that we denote by $\mathbf{t}_k$ and $\mathbf{f}_k$ respectively. The features can be both static or dynamic; however, the feature set needs to be consistent across all the events. For instance, the age and weight of a patient change across hospital visits, whereas gender can be assumed to remain the same. $\mathcal{T} \subset \mathcal{I}$ is the target set. The target set is different from the input set of events. For instance, in the case of hospital visits, the target set may consist of only diagnoses, whereas the set of all possible events can additionally contain procedures and treatments.

We use $\mathcal{M}$ to denote the model being trained for a given task. $\mathcal{M}$ has a module called an encoder, denoted by $\mathcal{M}_E$, for encoding the input sequences along with a given set of features corresponding to each event in the sequence. We also use $\mathcal{A}_E$ to denote an auxiliary encoder model as described in Section \ref{SubSection: Self-supervised Pre-training}. For instance, the auxiliary encoder $\mathcal{A}_E$ can be modeled using an affine layer, and $\mathcal{M}$ can be modeled using a transformer \citep{transformers}.

\subsection{Learning Contextual Embeddings of Events}
\label{SubSection: Self-supervised Pre-training}

\begin{figure*}[t]
    \centering
    \includegraphics[width=\linewidth]{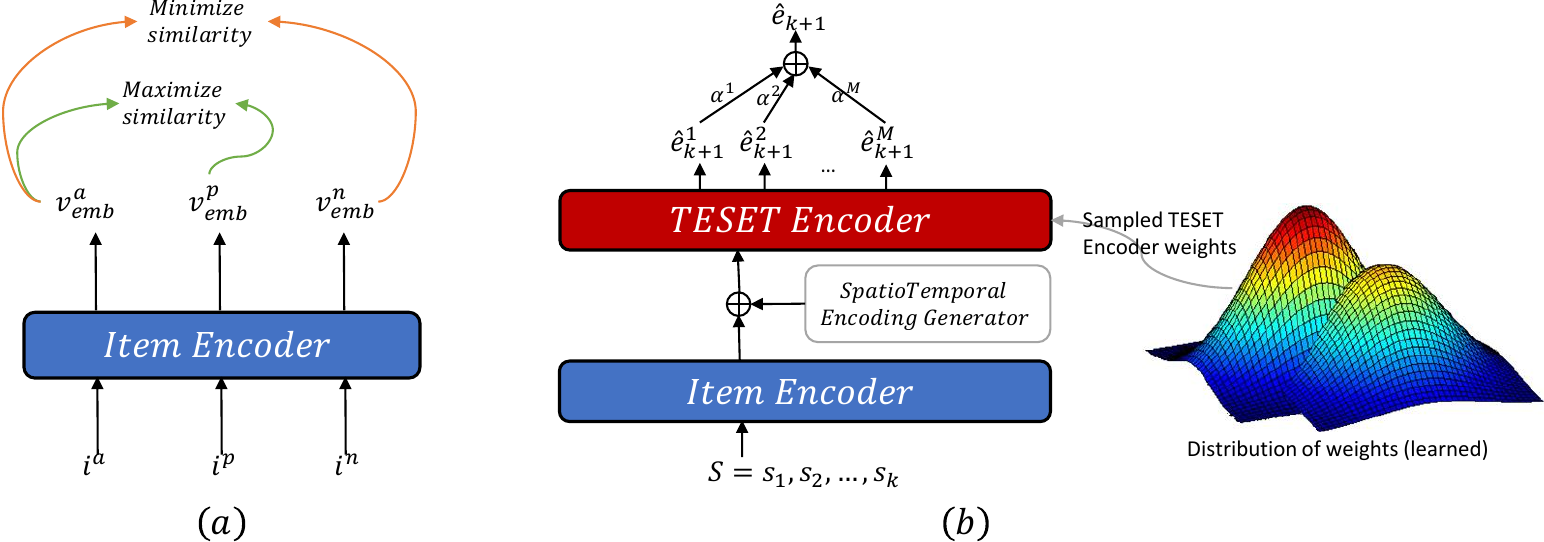}
    \caption{\small [Best viewed in color] Block diagram of our proposed approaches: (a) Learning event set representations, and (b) Inference procedure of our Bayesian Transformer based TESET model.}
    \label{Figure:OurApproachesBlockDiagram}
\end{figure*}

A measure of similarity among the vector representations of events $\mathbf{i} \in \mathcal{I}$ is required among the co-occurring events for learning meaningful contextual event embeddings. Consequently, in the first step of training, we use a self-supervised noise contrastive pre-training objective, similar to Noise Contrastive Estimation \citep{pmlr-v9-gutmann10a}, for learning vector representations corresponding to every event in the set $\mathcal{I}$. The input to $\mathcal{A}_E$ is an event $\mathbf{i} \in \mathcal{I}$ and the output of $\mathcal{A}_E$ is $\mathbf{v}_{emb}$, which is a $\mathbf{d}_{emb}$--dimensional embedding vector.

To train this encoder network, we iterate over each $\mathbf{s}_k$ in $\mathcal{S}$, for all sequences in the dataset. At each iteration, we have the event set $\mathbf{s}_k$, which consists of a set of events. We sample two events $\mathbf{i}^a$ and $\mathbf{i}^p$ uniformly at random from $\mathbf{s}_k$, which becomes our anchor sample and positive sample respectively. We similarly sample our negative sample $\mathbf{i}^n$ uniformly at random from among the events that are not present in the set $s_k$. i.e.
\begin{align}
    \mathbf{i}^a \sim \mathbb{U}(\mathbf{s}_k); 
    \mathbf{i}^p \sim \mathbb{U}(\mathbf{s}_k \backslash \{\mathbf{i}^a\}); 
    \mathbf{i}^n \sim \mathbb{U}(\mathcal{I} \backslash \mathbf{s}_k)
\end{align}
where $\mathbb{U}(\cdot)$ denotes sampling uniformly at random from a given set of events.
We then pass $\mathbf{i}^a$, $\mathbf{i}^p$, and $\mathbf{i}^n$ through $\mathcal{A}_E(\cdot)$ to obtain $\mathbf{v}_{emb}^a$, $\mathbf{v}_{emb}^p$, and $\mathbf{v}_{emb}^n$ respectively (as shown in Equation \ref{Equation: Item vector embedding}).

Then we calculate and maximize the following auxiliary contextual loss objective:
\begin{align}
    \label{Equation:Auxiliary Contextual Loss}
    \mathcal{L}_{aux} = \log(\sigma(\mathbf{v}_{emb}^a \cdot \mathbf{v}_{emb}^p)) + \log(1 - \sigma(\mathbf{v}_{emb}^a \cdot \mathbf{v}_{emb}^n))
\end{align}
where $\mathbf{a} \cdot \mathbf{b}$ represents the inner product among the vectors $\mathbf{a}$ and $\mathbf{b}$, and $\sigma(\mathbf{x})=1/1+e^{-\mathbf{x}}$ represents the sigmoid function.
Finally, the error is back-propagated through the auxiliary encoder model $\mathcal{A}_E$ and the corresponding parameters are updated using an appropriate optimizer.

\begin{algorithm}[t]
    \hrule
    \vspace{2pt}
    \caption{Contextual self-supervised representation learning of events in event set}
    \label{Algorithm: Phase1 training}
    \vspace{2.5pt}
    \hrule
    \vspace{3.5pt}
    \KwData{$\mathcal{D}$\Comment*[r]{Dataset}}
    \While{not converged}{
        $s_k \sim \mathcal{U}(\mathcal{D})$ \Comment*[r]{Sample event set from dataset}
        $i^a \sim \mathcal{U}(s_k), i^p \sim \mathcal{U}(s_k \backslash \{i^a\}), i^n \sim \mathcal{U}(\mathcal{I} \backslash s_k)$\Comment*[r]{Sample events}
        $v_{emb}^a = \mathcal{A}_E(i^a)$, where $i^a \sim \mathcal{U}(s_k)$ \Comment*[r]{Anchor}
        $v_{emb}^p = \mathcal{A}_E(i^p)$, where $i^p \sim \mathcal{U}(s_k \backslash \{i^a\})$ \Comment*[r]{Positive}
        $v_{emb}^n = \mathcal{A}_E(i^n)$, where $i^n \sim \mathcal{U}(\mathcal{I} \backslash s_k)$ \Comment*[r]{Negative}
        Update $\mathcal{A}_E$ using $\nabla \mathcal{L}_{aux} (v_{emb}^a,v_{emb}^p,v_{emb}^n)$ \Comment*[r]{Backpropagate}
    }
    \textbf{Return}: $\mathcal{A}_E$ \Comment*[r]{Return the trained encoder (embedding generator)}
    \vspace{1.5pt}
    \hrule
\end{algorithm}

\subsection{Temporal Event set (TESET) Modeling}
\label{SubSection: Hierarchical training}

After the auxiliary encoder model $\mathcal{A}_E$ is trained, it can generate embeddings as follows:
\begin{align}
    \label{Equation: Item vector embedding}
    \mathbf{v}_{emb} = \mathcal{A}_E(\mathbf{i})
\end{align}
In the next step of training, we train the encoder module $\mathcal{M}_E$ in our model $\mathcal{M}$.
For a given sequence $\mathcal{S}$, we assume the most recently occurred set of events to be $\mathbf{s}_k$. $\mathbf{s}_k$ also had associated $\mathbf{t}_k$ (a positive real value) as its corresponding timestamp, denoting when the event occurred in the timeline. Additionally $\mathbf{s}_k$ may also optionally contain an associated set of features $\mathbf{f}_k$.

All the previous set of events along with their corresponding timestamps and features until $\mathbf{s}_k$ is assumed to be history as follows:
\begin{align}
    \mathcal{H}_k = \{\langle \mathbf{s}_1, \mathbf{t}_1, \mathbf{f}_1 \rangle, \langle \mathbf{s}_2, \mathbf{t}_2, \mathbf{f}_2 \rangle, ..., \langle \mathbf{s}_{k-1}, \mathbf{t}_{k-1}, \mathbf{f}_{k-1} \rangle\}
\end{align}

We denote the target event set $\mathbf{e}_{k+1}$ as
\begin{align}
    \mathbf{e}_{k+1} = \mathbf{e}_{k+1}\cap\mathcal{T}
\end{align}
The objective in this step is to predict the tuple $\langle \mathbf{e}_{k+1}, \mathbf{t}_{k+1} \rangle$ given the tuple $\langle \mathbf{s}_k, \mathbf{t}_k, \mathbf{f}_k, \mathcal{H}_k \rangle$ as the input. In other words, the goal is to model the prediction of the set of next events along with the timestamp when it is supposed to occur given the most recent event, its timestamp, its associated features, and the entire history of events in that sequence of events.

Notice that the events $\mathbf{s}_k \in \mathcal{S}$ are sets of events. Hence, the events do not necessarily consist of only singleton elements and may contain two or more events. Consequently, we require $\mathcal{M}_E$ to be composed of a hierarchy of encoders: \textbf{(i)} Set Encoder that will combine the sets and give a single representation for all the events in the set, and \textbf{(ii)} Sequential Encoder that will take these combined representations as input and encode them temporally.

However, this approach possesses its own set of challenges as follows:
\begin{compactitem}
    \item [(a)] Set encoding is a difficult problem since the set representations must satisfy properties such as permutation invariance and equivariance.
    \item [(b)] During implementation, the event encoder either requires to be duplicated with one copy corresponding to every event set $s_1,...,s_k$ or techniques such as gradient accumulation are needed while training. The duplication again requires high accelerator memory and efficient coding to utilize parallelization properly.
    \item [(c)] The sequential nature of the event encoder prevents efficient parallelization, on top of it waiting for the event encoder to get the set representations.
\end{compactitem}
\vspace{0.08in}
As a solution to these problems, we propose a transformer-based architecture for training the model's encoder module $\mathcal{M}_E$.
We stack all the events in the most recent occurred event $\mathbf{s}_k$ together with all the events $\mathbf{s}_1, ..., \mathbf{s}_{k-1}$ in history $\mathcal{H}_k$. Thus, assuming the events in each event set $\mathbf{s}_j$ are: $\mathbf{i}_j^1, \mathbf{i}_j^2, ..., \mathbf{i}_j^{|\mathbf{s}_j|}$, the current event set along with the history event sets in a given sequence $\mathcal{S}$ becomes
\begin{align}
    \mathcal{S}_k = \mathbf{i}_1^1, \mathbf{i}_1^2, ..., \mathbf{i}_1^{|\mathbf{s}_1|}, \mathbf{i}_2^1, \mathbf{i}_2^2, ..., \mathbf{i}_2^{|\mathbf{s}_2|}, ..., \mathbf{i}_{k}^1, \mathbf{i}_{k}^2, ..., \mathbf{i}_{k}^{|\mathbf{s}_k|}
\end{align}

\begin{algorithm}[t]
    \hrule
    \vspace{2pt}
    \caption{Representation learning of temporal event sets}
    \label{Algorithm: Phase2 training}
    \vspace{2.5pt}
    \hrule
    \vspace{3.5pt}
    \KwData{$\mathcal{D}$\Comment*[r]{Dataset}}
    \While{not converged}{
        $\mathcal{S} \sim \mathcal{U}(\mathcal{D})$ \Comment*[r]{Sample a sequence from dataset}
        $k \sim \mathcal{U}([|S|])$ \Comment*[r]{Sample an integer $1 \leq k \leq len(\mathcal{S})$}
        $ \langle \hat{\mathbf{e}}_{k+1}, \hat{\mathbf{t}}_{k+1} \rangle = \mathcal{M}_E ( \langle s_k, t_k, f_k, \mathcal{H}_k \rangle )$ \Comment*[r]{Forward pass}
        Update $\mathcal{M}_E$ and $\mathcal{A}_E$ using $\nabla\mathcal{L}(\langle \hat{\mathbf{e}}_{k+1}, \hat{\mathbf{t}}_{k+1} \rangle,\langle \mathbf{e}_{k+1}, \mathbf{t}_{k+1} \rangle)$ \Comment*[r]{Backpropagate}
    }
    \textbf{Return}: $\mathcal{A}_E$ and $\mathcal{M}_E$ \Comment*[r]{Return the learned representation models}
    \vspace{1.5pt}
    \hrule
\end{algorithm}

In order to differentiate among the various event sets, we use the following techniques that are specifically applicable to a transformer-based architecture: (i) Special Tokens, and (ii) Custom SpatioTemporal Encodings containing both positional and temporal information.

\paragraph{Special Tokens:} We use a special token, which is often referred to as the separator token (denoted by \textmyfont{[SEP]}) in the literature, after the event listed as $\mathbf{i}_j^{|\mathbf{s}_j|}$ for all $j \leq k$. This enables us to separate the event sets from each other whilst also providing us with a representation corresponding to each event set in the sequence $\mathcal{S}_k$. We use another classifier special token (denoted by \textmyfont{[CLS]}) at the very end of the sequence $\mathcal{S}_k$. This token helps us summarize the contents of the entire sequence, and its corresponding vector can be used for downstream tasks. We denote the resultant augmented sequence of events and tokens as $\mathcal{S}_k^*$. In the rest of this section, we will assume all the elements in the sequence $\mathcal{S}_k^*$ to be tokens to keep the discussion and notations uniform with the transformer literature. Hence the augmented sequence becomes
\begin{align}
    \mathcal{S}_k^* = \mathbf{i}_1^1, ..., \mathbf{i}_1^{|\mathbf{s}_1|}, \textmyfont{[SEP]}, \mathbf{i}_2^1, ..., \mathbf{i}_2^{|\mathbf{s}_1|}, \textmyfont{[SEP]}, ..., \textmyfont{[SEP]}, \mathbf{i}_{k}^1,  ..., \mathbf{i}_{k}^{|\mathbf{s}_k|}, \textmyfont{[SEP]}, \textmyfont{[CLS]}
\end{align}

\paragraph{SpatioTemporal Embeddings:} Next, we add custom Spatial and Temporal (SpatioTemporal) Encodings to all the events in the augmented sequence $\mathcal{S}_k^*$. The transformer framework assumes the entire sequence $\mathcal{S}_k$ as an atomic unit. It is therefore essential that we specify an encoding vector for each token in $\mathbf{s}_k$ such that it can not only enable the model to differentiate among various event sets in the sequence but also effectively accumulate contextual information in the respective output embeddings. Our requirement for encoding is different from positional encoding \citep{transformers} due to the following reasons: \textbf{(i)} events $\mathbf{i}_{j}^1, \mathbf{i}_{j}^2, ..., \mathbf{i}_{j}^{|\mathbf{s}_j|}$ within a set $\mathbf{s}_j$ for all $j \leq k$ are unordered, unlike the ordered words in a textual sequence, and \textbf{(ii)} two consecutive timesteps are not uniformly separated in the timeline. For instance, the duration between the current and next visit of a patient might vary from as short as a month to as long as multiple years.

Consequently, we use the SpatioTemporal Encodings as described below, which can handle these non-uniform temporal differences whilst also retaining information about the co-occurrence of events in event sets.
\begin{align}
    \mathbf{v}_{enc}^{pos}(j, d) &= \begin{cases}
        \sin{(j/10000^\frac{2d}{\mathbf{d}_{emb}})} & ; \texttt{if } j \texttt{ is even}\\
        \cos{(j/10000^\frac{2d}{\mathbf{d}_{emb}})} & ; \texttt{otherwise}
    \end{cases}\notag
    \\
    \mathbf{v}_{enc}^{temp}(\mathbf{t}_j, d) &= \begin{cases}
        \sin{(\mathbf{t}_j/10000^\frac{2d}{\mathbf{d}_{emb}})} & ; \texttt{if } \mathbf{t}_j \texttt{ is even}\\
        \cos{(\mathbf{t}_j/10000^\frac{2d}{\mathbf{d}_{emb}})} & ; \texttt{otherwise}
    \end{cases}\notag
    \\
    \mathbf{v}_{enc}(j, \mathbf{t}_j, d) &= \mathbf{v}_{enc}^{pos}(j, d) + \mathbf{v}_{enc}^{temp}(\mathbf{t}_j, d)
\end{align}
where $\mathbf{t}_j$ is the timestamp corresponding to the $j^{th}$ event set $\mathbf{s}_j$ for $1 \leq j \leq k$.

Note that the initial value of $\mathbf{v}_{emb}$ is obtained by passing each event from event sets through $\mathcal{A}_E$. Then we add $\mathbf{v}_{enc}$ to $\mathbf{v}_{emb}$ before passing it on to the Transformer model. We get a $\mathbf{d}_{emb}$--dimensional embedding vector $\mathbf{v}_{emb}^{\textmyfont{[CLS]}}$ corresponding to the \textmyfont{[CLS]} token. We denote this output vector by $\mathbf{v}_{out}$. Refer to Figure \ref{Figure:OurApproachesBlockDiagram} for a block diagram of our approach.

Additionally, we use the following two prediction heads for training the representation model $\mathcal{M}$: \textbf{(i)} Event set Prediction Head, denoted by $\mathcal{P}_E$, and \textbf{(ii)} Temporal Prediction Head, denoted by $\mathcal{P}_T$. $\mathcal{P}_E$ takes $\mathbf{v}_{out}$ as input and predicts $M$ pairs of Gaussian distributional parameter vectors
$\langle\mu_{\hat{\mathbf{e}}_{k+1}}^1,\sigma_{\hat{\mathbf{e}}_{k+1}}^1\rangle,$
$\langle\mu_{\hat{\mathbf{e}}_{k+1}}^2,\sigma_{\hat{\mathbf{e}}_{k+1}}^2\rangle,$
$...,\langle\mu_{\hat{\mathbf{e}}_{k+1}}^M,\sigma_{\hat{\mathbf{e}}_{k+1}}^M\rangle$ along with $M$ mixing coefficients $\alpha_{\hat{\mathbf{e}}_{k+1}}^1,\alpha_{\hat{\mathbf{e}}_{k+1}}^2,...,\alpha_{\hat{\mathbf{e}}_{k+1}}^M$. The $M$ event set prediction vectors $\hat{\mathbf{e}}_{k+1}^1,\hat{\mathbf{e}}_{k+1}^2,...,\hat{\mathbf{e}}_{k+1}^M$ are sampled from the distribution parameters with the same number of dimensions as events in the target set $\mathcal{T}$ with each dimension modeling a Bernoulli distribution. After mixing the sampled vectors according to the mixing coefficients, we get
\begin{align}
    \label{Equation:Mixing}
    \hat{\mathbf{e}}_{k+1} = \alpha_{\hat{\mathbf{e}}_{k+1}}^1\cdot\hat{\mathbf{e}}_{k+1}^1,\alpha_{\hat{\mathbf{e}}_{k+1}}^2\cdot\hat{\mathbf{e}}_{k+1}^2,...,\alpha_{\hat{\mathbf{e}}_{k+1}}^M\cdot\hat{\mathbf{e}}_{k+1}^M
\end{align}
Similarly, $\mathcal{P}_T$ takes $v_{out}$ as input and outputs the $M$ temporal Gaussian distributional parameter pairs
$\langle\mu_{\hat{\mathbf{t}}_{k+1}}^1,\sigma_{\hat{\mathbf{t}}_{k+1}}^1\rangle,$
$\langle\mu_{\hat{\mathbf{t}}_{k+1}}^2,\sigma_{\hat{\mathbf{t}}_{k+1}}^2\rangle,$
$...,\langle\mu_{\hat{\mathbf{t}}_{k+1}}^M,\sigma_{\hat{\mathbf{t}}_{k+1}}^M\rangle$ along with $M$ temporal mixing coefficients $\alpha_{\hat{\mathbf{t}}_{k+1}}^1,\alpha_{\hat{\mathbf{t}}_{k+1}}^2,...,\alpha_{\hat{\mathbf{t}}_{k+1}}^M$. We sample scalars $\hat{\mathbf{t}}_{k+1}^1,\hat{\mathbf{t}}_{k+1}^2,...,\hat{\mathbf{t}}_{k+1}^M$ and similar to Equation \ref{Equation:Mixing}, we obtain $\hat{\mathbf{t}}_{k+1}$ by mixing them according to the mixing coefficients.
We use reparametrization (similar to \cite{vae}) to enable backpropagation in our model.

\paragraph{Temporal Event set Modeling:}In order to learn representations in the second step of pre-training, we propose the Temporal Event set Modeling objective as described below.
Upon sampling the tuple $\langle \hat{\mathbf{e}}_{k+1}, \hat{\mathbf{t}}_{k+1} \rangle$ corresponding to an input $\langle \mathbf{s}_k, \mathbf{t}_k, \mathbf{f}_k, \mathcal{H}_k \rangle$, we use the following loss objectives as a part of our TESET Modeling:
\begin{compactitem}
    \item[\textbf{(i)}] An element-wise binary cross-entropy loss on $\hat{\mathbf{e}}_{k+1}$ against $\mathbf{e}_{k+1}$:
    \begin{align}
        \mathcal{L}_{Event}^{BCE} &= \frac{1}{|\mathcal{T}|} \sum_{d\in[|\mathcal{T}|]} \mathbbm{1}_{\{\mathcal{T}^{(d)}\in \mathbf{e}_{k+1}\}}\hat{\mathbf{e}}_{k+1}^{(d)} + \mathbbm{1}_{\{\mathcal{T}^{(d)}\notin \mathbf{e}_{k+1}\}}(1-\hat{\mathbf{e}}_{k+1}^{(d)})
    \end{align}
    where $\mathbbm{1}$ denotes the indicator function and $\mathbf{v}^{(d)}$ represents the $d^{th}$ dimension of the vector $\mathbf{v}$.

\begin{table*}[!t]
    \centering
    \caption{\small \textbf{Temporal Event set Modeling Results.} We compare our approaches to baselines. For DSC, the larger the better; for MAE, the smaller the better. We can see that even without Contextual embeddings, our methods outperform the baselines. Best results are in bold}
    \label{Table: Temporal Event Modeling Results}
    \begin{tabular}{lcccc}
        \toprule
        \multicolumn{1}{c}{\multirow{2}{*}{Training method}} & \multicolumn{2}{c}{Synthea} & \multicolumn{2}{c}{Instacart} \\ \cline{2-5} 
        \multicolumn{1}{c}{} & \begin{tabular}[c]{@{}c@{}}Event set\\pred (DSC)\end{tabular} & \begin{tabular}[c]{@{}c@{}}Time pred\\ (MAE)\end{tabular} & \begin{tabular}[c]{@{}c@{}}Event set\\pred (DSC)\end{tabular} & \begin{tabular}[c]{@{}c@{}}Time pred\\ (MAE)\end{tabular} \\
        \midrule
        \textit{Baselines:} &  &  &  &  \\
        Neural Hawkes Process & 0.08  & 2.50 & 0.29 & 0.24 \\
        Transformer Hawkes Process & 0.18 & 2.41 & 0.32  & 0.24 \\
        Hierarchical Model & 0.12 & 2.51 & 0.30 & 0.23 \\
        \midrule
        \textit{Ours:} &  &  &  &  \\
        TESET & 0.20 & 2.29 & 0.35 & 0.21 \\
        TESET + Contextual Embeddings & \textbf{0.30} & \textbf{2.17} & \textbf{0.42} & \textbf{0.18} \\
        \bottomrule
    \end{tabular}
\end{table*}
    
    \item[\textbf{(ii)}] Additionally, we use dice loss for handling the class imbalance problem in $s_{k+1}\cap\mathcal{T}$.
    \begin{align}
        \mathcal{L}_{Event}^{Dice} = 1 - \frac{1}{|\mathcal{T}|}\sum_{d\in[|\mathcal{T}|]} \frac{2\text{ }\hat{\mathbf{e}}_{k+1}^{(d)}\text{ }\mathbf{e}_{k+1}^{(d)}+\epsilon}{\sum_{d'\in[|\mathcal{T}|]} \hat{\mathbf{e}}_{k+1}^{(d')} + \mathbf{e}_{k+1}^{(d')}+\epsilon}
    \end{align}
    where $\epsilon$ is a small Laplace Smoothening constant.
    
    \item[\textbf{(iii)}] Finally, for the timestamp prediction head, we use Huber loss to align $\hat{\mathbf{t}}_{k+1}$ and $\mathbf{t}_{k+1}$:
    \begin{align}
        \mathcal{L}_{Temporal}^{Huber} &=
        \begin{cases}
            \Delta^2/2 & ; \text{if } \Delta < \delta\\
            \delta(\Delta-\delta/2) & ; \text{otherwise} 
        \end{cases}&&
    \end{align}
    where $\Delta$ is the absolute value of $\hat{\mathbf{t}}_{k+1} - \mathbf{t}_{k+1}$
    and $\delta$ is a positive constant.
\end{compactitem}
We minimize a linear combination of the above loss objectives as follows:
\begin{align}
    \label{Equation: Phase II Loss}
    \mathcal{L} = \lambda_1\mathcal{L}_{Event}^{BCE} + \lambda_2\mathcal{L}_{Event}^{Dice} + \lambda_3\mathcal{L}_{Temporal}^{Huber}
\end{align}
where, $\lambda_1, \lambda_2, \lambda_3 > 0$. We calculate this loss and back-propagate the gradients through our encoder models $\mathcal{M}_E$ and $\mathcal{A}_E$ and update the model parameters accordingly. \\

\noindent \textbf{Multiple Generations}: We implement our Transformer model $\mathcal{M}$ using the Probabilistic Bayesian neural network framework. Essentially, every time we train the model we sample the weights and biases (for every layer) from a weight distribution, and then update the distribution through backpropagation. This enables us to sample the weights multiple times, thus providing us with an ensemble of $N$ networks whose predictions we combine to get our final predicted outputs. This stabilizes the training, helps converge the loss objective faster, and the validation metrics improve noticeably during the initial stages of training (see Section \ref{Section: Results} for more details).

\section{Experiments\text{\normalfont \whatb{$^{2}$}}}

\footnotetext[2]{Codes for our experiments are available at: \href{https://github.com/paragduttaiisc/temporal_event_set_modeling}{https://github.com/paragduttaiisc/temporal\_event\_set\_modeling}}

\subsection{Datasets}
\label{Subsection: Datasets}

\begin{table*}[!t]
    \centering
    \caption{\small \textbf{Fine-tuning results.} FT stands `fine-tuned'. Our models consistently outperform the baselines in the setting of being fine-tuned for the downstream tasks. It should be noted that Fine tuning doesn't work on the baseline models, and they often perform worse. In each stratum, the best-performing models have been italicized. The best-performing models have been shown in bold.}
    \label{Table: Fine-tuning}
    \begin{tabular}{clcccc}
        \toprule
        \multirow{2}{*}{FT?} & \multicolumn{1}{c}{\multirow{2}{*}{Training method}} & \multicolumn{2}{c}{Synthea} & \multicolumn{2}{c}{Instacart} \\ \cline{3-6} 
        & \multicolumn{1}{c}{} & \begin{tabular}[c]{@{}c@{}}Event set\\ given time\\ (DSC)\end{tabular} & \begin{tabular}[c]{@{}c@{}}Time\\ given event\\ (MAE)\end{tabular} & \begin{tabular}[c]{@{}c@{}}Event set\\ given time\\ (DSC)\end{tabular} & \begin{tabular}[c]{@{}c@{}}Time\\ given event\\ (MAE)\end{tabular} \\
        \midrule
        \multirow{4}{*}{\shortstack[c]{Trained \\from \\scratch}} & Neural Hawkes Process & 0.21 & 5.70 & 0.35 & 2.19 \\
        & Transformer Hawkes Process & 0.20 & 4.52 & 0.34 & 2.15 \\
        & Hierarchical Model & 0.19 & 5.29 & 0.34 & 2.20 \\
        & TESET (Ours) & \textit{0.22} & \textit{4.28} & \textit{0.38} & \textit{1.83} \\
        \midrule
        \multirow{4}{*}{\begin{tabular}[c]{@{}c@{}}Fine-\\tuned\end{tabular}} & Neural Hawkes Process & 0.13 & 6.01 & 0.30  & 2.29 \\
        & Transformer Hawkes Process & 0.19  & 4.60 & 0.33 & 2.24 \\
        & Hierarchical Model & 0.18 & 5.87 & 0.35 & 2.31 \\
        & TESET (Ours) & \textit{\textbf{0.25}} & \textit{\textbf{3.91}} & \textit{\textbf{0.41}} & \textit{\textbf{1.19}} \\
        \bottomrule
    \end{tabular}
\end{table*}


\begin{compactitem}
    \item [1.] \textbf{Synthea:}
    \citep{synthea} encompasses the comprehensive medical records of each patient generated synthetically. The medical history of each patient is represented as a sequential sequence of their hospital visits, along with the corresponding timestamps denoting the time of each visit. Each hospital visit comprises an event set, containing the diagnoses, treatments, and procedures administered during that particular visit, in addition to the patient's characteristics such as age, weight, and gender.
    \item [2.] \textbf{Instacart:}
    \citep{instacart} is a comprehensive collection of customers' order histories. Each individual customer's order history is represented as a sequential arrangement of orders, wherein each order includes a set of items purchased by the respective customer and the corresponding timestamp indicating the time of the order.
    \item [3.] \textbf{MIMIC-III:}
    \citep{mimic} provided includes the historical data of patients who have visited the Intensive Care Unit (ICU) at a hospital. By analyzing the Electronic Health Record (EHR) history of each patient, we extract the sequential information regarding the set of medical conditions diagnosed and the respective admission timestamps. For the purpose of the finetuning task, the medical codes used in the Synthea dataset are correspondingly mapped to the medical codes employed in the MIMIC-III dataset.
\end{compactitem}

\subsection{Baselines}
We quantify the advantage of our proposed approach by comparing with the following competitive baselines\footnote[3]{Baselines 1 and 2 models are originally used for event prediction given a sequence of events and Baseline 3 is originally used in the discrete timestep set prediction. We extend them to predict sets in the continuous domain.}:
\begin{wrapfigure}{r}{0.5\textwidth}
  \begin{center}
    \centering
    \includegraphics[width=0.49\textwidth]{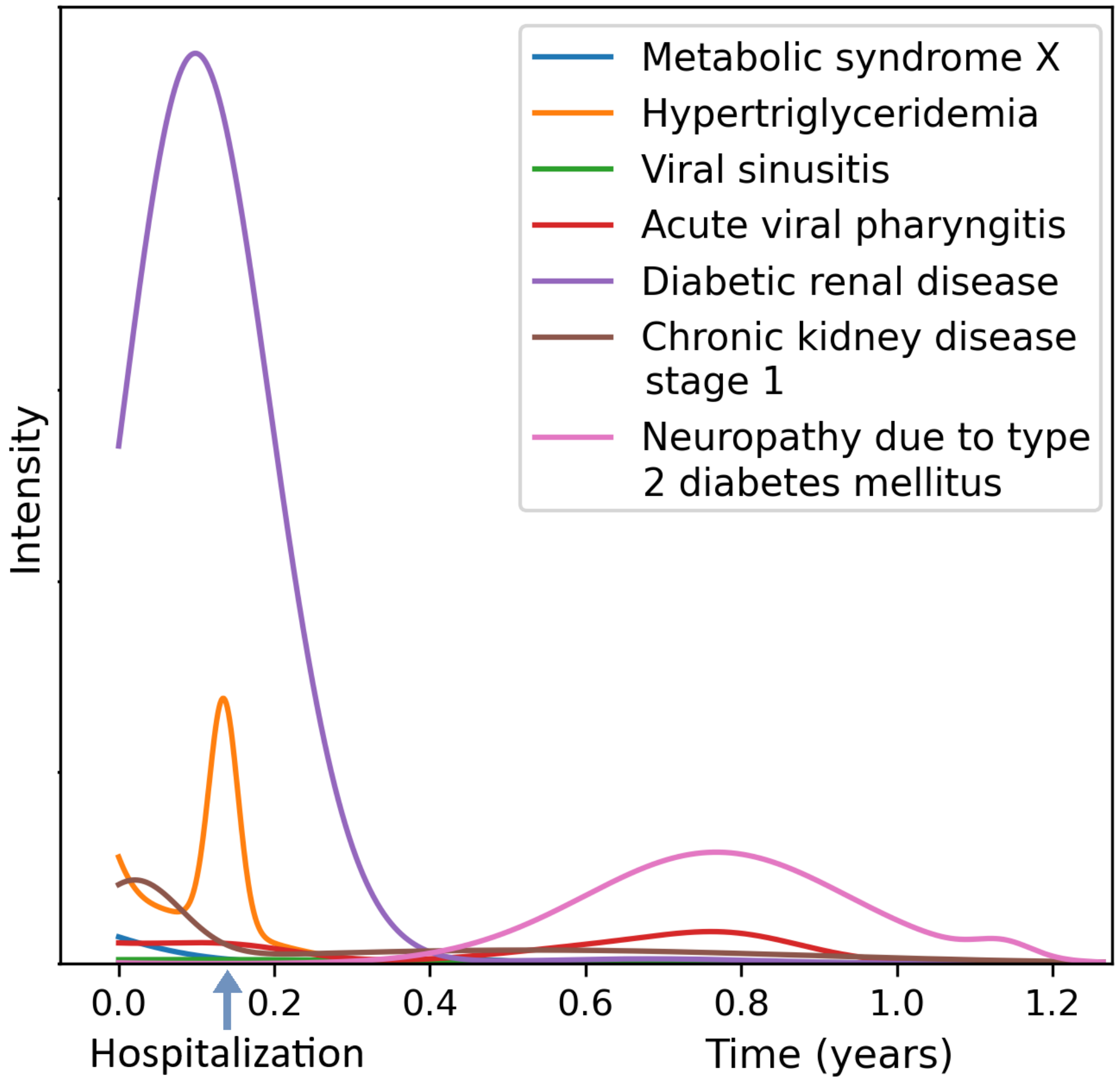}
    \caption{\small [Best viewed in color] Our model's predicted intensity plot for an elderly female patient who had a history of diabetes. The peak of two high-intensity disease curves (Diabetic Renal Disease and Hypertriglyceridemia) coincides with the date of actual hospitalization. Also, it is predicted that Neuropathy might be a problem in the future, which is a well-known condition for people suffering from Type II Diabetes.}
    \label{Figure: IntensityAblation}
  \end{center}
\end{wrapfigure}

\begin{compactitem}
    \item[1.] \textbf{Neural Hawkes Process (NHP):} The NHP \citep{nhp} employs a Recurrent Neural Network, specifically a continuous-LSTM, to parameterize the intensity function $\lambda$ of the Hawkes process. The intensity function is $K$-dimensional, denoted as $\lambda_k(t) = f_k(w_k^Th(t)$; where, $f_k(.)$ is the decay function $\delta$ that is chosen to be the softplus function, $K$ is the number of events, and $h(t)$ is the hidden state of the LSTM.
\end{compactitem}

\begin{table}[t]
    \centering
    \caption{\small \textbf{Transfer learning results}. We fine-tune the TEM model trained on the synthetic dataset using the MIMIC-III dataset. It is observable that syntheic to real transfer works better for our approach.}
    \label{Table: Transfer Learning}
    \begin{tabular}{lcccc}
        \toprule
        \multicolumn{1}{c}{Model} & \begin{tabular}[c]{@{}c@{}}Event set\\ (DSC)\end{tabular} & \begin{tabular}[c]{@{}c@{}}Time\\ (MAE)\end{tabular} & \begin{tabular}[c]{@{}c@{}}Event set given\\ time (DSC)\end{tabular} & \begin{tabular}[c]{@{}c@{}}Time given\\ event (MAE)\end{tabular} \\
        \midrule
        TESET trained from scratch & 0.49 & 0.67 & 0.47 & 0.70 \\
        TESET fine-tuned & \textit{0.52} & \textit{0.14} & \textit{0.50} & \textit{0.19} \\
        \bottomrule
    \end{tabular}
\end{table}

\begin{compactitem}
    \item[2.] \textbf{Transformer Hawkes Process (THP):} The THP \citep{thawkes} utilizes a self-attention mechanism and temporal encoding to model the Hawkes process. This approach effectively captures long-term dependencies while maintaining computational efficiency, distinguishing it from the NHP (Neural Hawkes Process).
\end{compactitem}
    
\begin{compactitem}
    \item[3.] \textbf{Hierarchical Model (HM):} HM uses a hierarchical encoder where in the first step it encodes the sets and provides a representation for each set using a pooling function and in the second step it encodes the set\\representations temporally. We use a fully\\connected neural network to encode the\\sets and a Bi-LSTM model for encoding the set representations temporally. This is in similar lines to the Sets2Sets model \citep{sets2sets}.
\end{compactitem}

\subsection{Downstream tasks}

\begin{table}[t]
    \centering
    \caption{\small \textbf{SpatioTemporal Encodings}: Need for custom encoding during TEM is evident from the considerable advantage we observe. Models were trained on the Instacart dataset.}
    \label{Table: Custom encoding ablation}
    \begin{tabular}{lcc}
        \toprule
        Transformer Encoding & Event set pred. (DSC) & Time pred. (MAE) \\
        \midrule
        Positional Enc \citep{transformers} & 0.35 & 0.22 \\
        SpatioTemporal Enc (Ours) & \textbf{0.42} & \textbf{0.18} \\
        \bottomrule
    \end{tabular}
\end{table}

\begin{wrapfigure}{r}{0.50\textwidth}
  \begin{center}
    \includegraphics[width=0.49\textwidth]{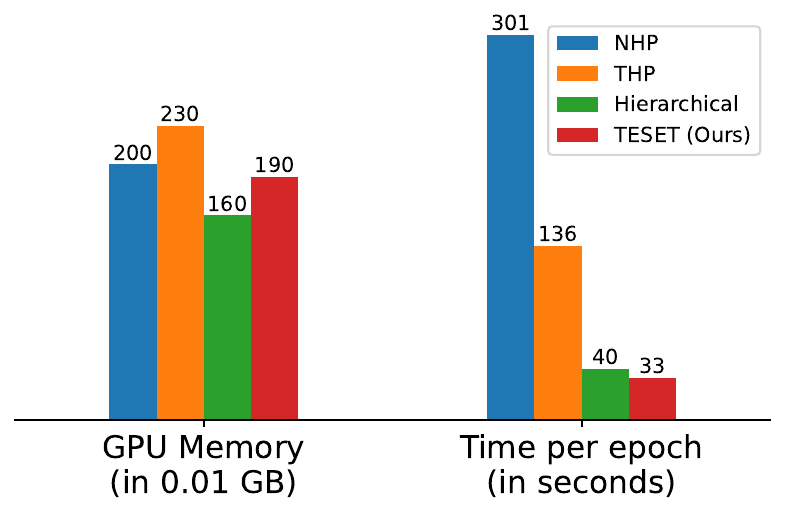}
  \end{center}
  \caption{\small [Best viewed in color] Resource usage and training time comparison during TEM training on Synthea dataset. The TESET model is the fastest although it has similar computational requirements.}
  \label{Figure: Training and Inference Time comparison}
\end{wrapfigure}

We demonstrate the superiority of the representations learned by our TESET model by fine-tuning on the following downstream tasks:

\begin{compactitem}
    \item[1.] \textbf{Event set prediction given time:}
    In this downstream task, the idea is to model $\mathbf{s}_{k+1}$ given the tuple $(\mathbf{s}_k, \mathbf{t}_k, \mathbf{f}_k,$
    $\mathcal{H}_k, \mathbf{t}_{k+1})$. In other words, we would like to predict the event set that might occur in the future given the future timestamp in addition to the tuple of the most recent event, its timestamp, its associated features, and the entire history of events-sets in that sequence as mentioned in Section \ref{Section: Proposed Approaches}. Note that the future timestamp when we want to predict the most probable event set lies in the continuous domain.
\end{compactitem}

\begin{compactitem}
    \item[2.] \textbf{Temporal prediction given an event:}
    Conversely, in this downstream task, the idea is to model $\mathbf{t}_{k+1}$ given the tuple $(\mathbf{s}_k, \mathbf{t}_k, \mathbf{f}_k, \mathcal{H}_k, \mathbf{i})$, where $\mathbf{i} \in \mathcal{T}$. In other words, we would like to predict the most probable time when a particular event might occur in the future given the event from the target set in addition to the tuple of a most recent event, its timestamp, its associated features, and the entire history of events in that sequence of events.    
\end{compactitem}

\subsection{Ablation Studies}
We formulate our ablation experiments in the form of the following interesting research questions:\\
\noindent\textbf{RQ-1:} Are the representations learned by TEM useful for related tasks?\\
\noindent\textbf{RQ-2:} What is the role of incorporating additional features into our framework?\\
\noindent\textbf{RQ-3:} How effective is the Contextual Event Representation Learning step?\\
\noindent\textbf{RQ-4:} How likely are our approaches to adapt and generalize with respect to a domain shift?\\
\noindent\textbf{RQ-5:} Can the advantage of using SpatioTemporal encodings over conventional positional encodings be quantified?\\
\noindent\textbf{RQ-6:} What is the training time saved by considering the set of temporal data points rather than each event individually?\\
\noindent\textbf{RQ-7:} Is the Bayesian Transformer with the distributional heads even required?\\
\noindent\textbf{RQ-8:} Do the predicted event intensities for a given history sequence correspond to something meaningful?



\section{Results}
\label{Section: Results}



\begin{figure}[t]
    \centering
    \begin{minipage}{.45\textwidth}
        \centering
        \includegraphics[width=\linewidth]{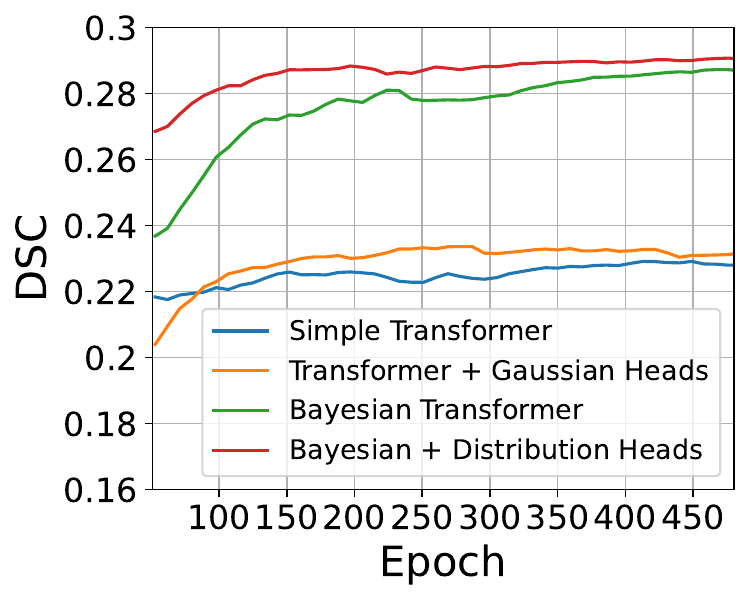}
    \end{minipage}%
    \begin{minipage}{.5\textwidth}
        \centering
        \includegraphics[width=\linewidth]{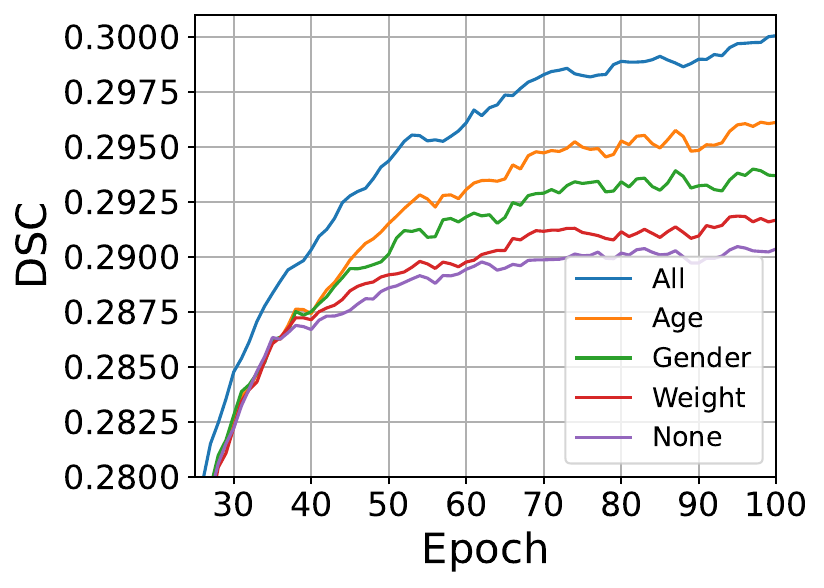}
    \end{minipage}
    \caption{\small [Best viewed in color] The plot on the \textbf{left} compares the test set dice scores for the TESET variants. Bayesian Transformer extends Simple Transformer with the Probabilistic Bayesian NN framework, while the Transformer with Gaussian Heads predict single Gaussian distribution at the final layer. It is clearly observable that Bayesian Transformer with Distributional Heads (Ours) is more stable and performs well right from the start. The plot on the \textbf{right} plots the test set dice scores of the TESET model with a combination of various features. It can be noticed that using all the features has considerable advantage.}
    \label{Figure: Features Ablation}
\end{figure}

Table \ref{Table: Temporal Event Modeling Results} compares the performance of our proposed models with baselines under similar sett-\\ings on two different tasks and two datasets:
\begin{compactitem}
    \item [(i)] It is evident that our TESET model outperforms existing baselines in both event set and temporal prediction metrics. We achieve 0.12 and 0.10 DSC improvement (absolute metrics) in Synthea and Instamart datasets respectively for the event set prediction sub-task. We also achieve 0.34 and 0.05 absolute improvement in MAE in the same datasets for the time prediction sub-task.
\end{compactitem}

\begin{compactitem}
    \item [(ii)] We can quantify \textbf{RQ-3} by looking at the difference of metrics with and without using contextual embeddings. It can be noticed that using contextual embedding is clearly advantageous.
\end{compactitem}

\begin{wrapfigure}{r}{0.61\textwidth}
  \begin{center}
    \includegraphics[width=0.6\textwidth]{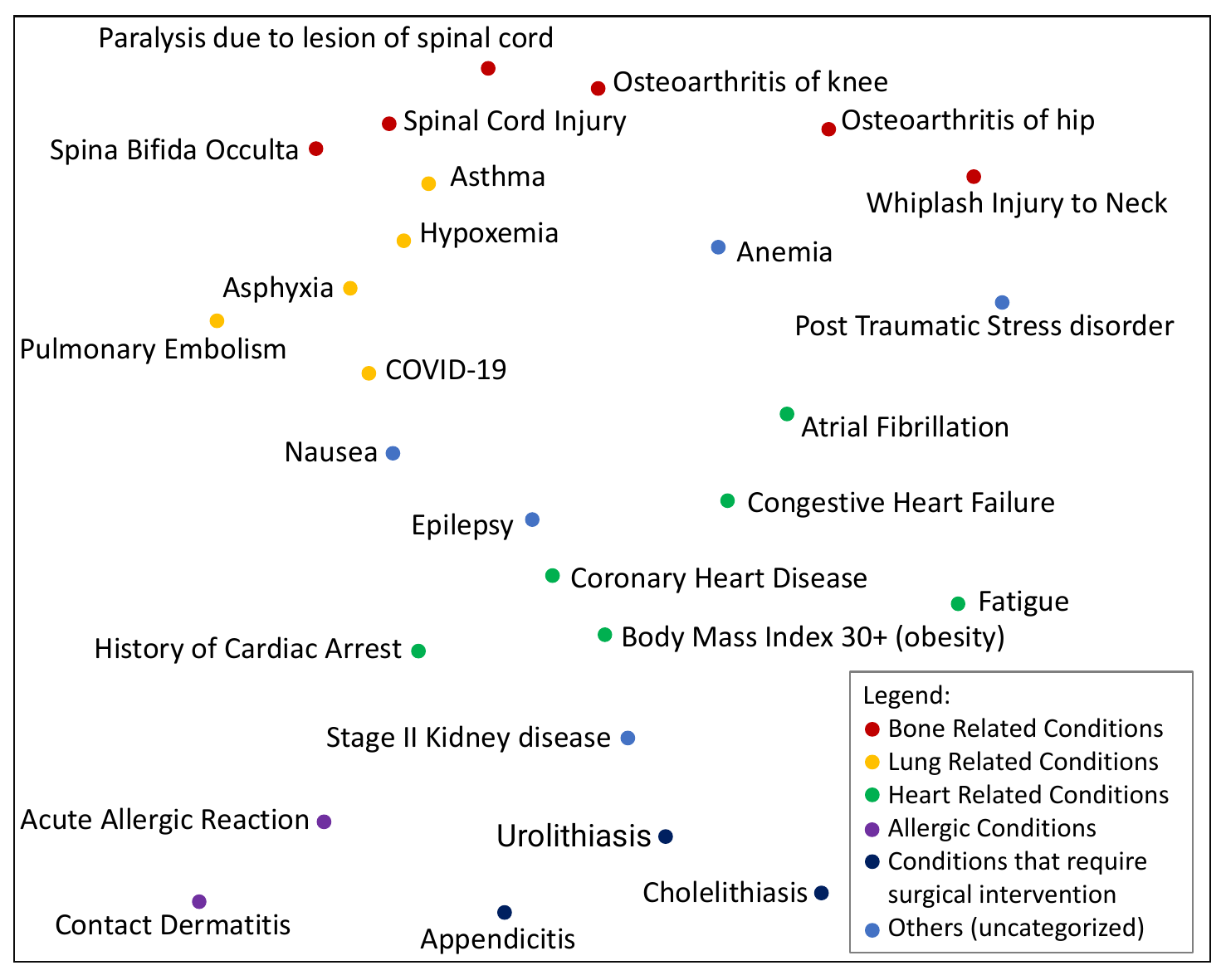}
  \end{center}
  \caption{\small [Best viewed in color] 2D t-SNE embeddings of the representations learned after first step of our approach in Synthea Dataset. It can be observed that clusters are formed in the embedding space.}
  \label{Figure: DiseaseTsne}
\end{wrapfigure}

Figure \ref{Figure: DiseaseTsne} additionally shows a magnified t-SNE plot of the Contextual vectors in the representation space to demonstrate the clustering of similar items.

Table \ref{Table: Fine-tuning} compares the fine-tuning results for the event set given time and time given event downstream task:
\begin{compactitem}
    \item [(i)] It can be noticed from the table that our methods outperform the baselines when fine-tuned instead of being trained from scratch.
    \item [(ii)] It is again evident that our model (TESET) learns representations during TEM that can be used for downstream tasks, thus answering \textbf{RQ-1}. On the other hand, the baseline approaches learn representations that are not generalizable for downstream tasks, and hence they perform better (compared to themselves) when trained from scratch.
\end{compactitem}

Figure \ref{Figure: Features Ablation} answers \textbf{RQ-2}. We can attribute the consistently improved performance of the model throughout the TEM training to the use of domain-specific features such as age, weight and gender. When trained with all of the features together, the model achieves the best performance.

\begin{table}
    \centering
    \caption{\small \textbf{Time and Computational Complexity.} An analysis of the computational and time complexity for each layer of the baseline methods and our method. The notations are as follows: $T$ indicates the Sequence Length, $\mu_E$ indicates the Average Event-Set Length (average number of items in the event-sets), and $d$ indicates the Embedding (hidden) dimension.}
    \begin{tabular}{lcccc}
        \toprule
         & NHP & THP & HM & TESET (Ours) \\
        \midrule
        Computational Complexity & $\bigO(T \cdot \mu_E \cdot d^2)$ & $\bigO(T^2 \cdot \mu_E^2 \cdot d)$ & $\bigO(T \cdot d)$ & $\bigO(T^2 \cdot d)$ \\
        Time Complexity & $\bigO(T \cdot \mu_E)$ & $\bigO(1)$ & $\bigO(1)$ & $\bigO(1)$ \\
        \bottomrule
    \end{tabular}
    \label{Table: Operations Complexity}
\end{table}

Table \ref{Table: Transfer Learning} presents the domain generalization capabilities of the representations learned by our TEM model, answering \textbf{RQ-4}. We can see that even though our TEM model was trained on the\\Synthea dataset, it generalizes quite effectively\\to the real-world dataset MIMIC-III.

Table \ref{Table: Custom encoding ablation} answers \textbf{RQ-5} by showing that our SpatioTemporal Encodings definitely score higher metrics in both event set prediction and time prediction during TEM when compared to vanilla Positional Embeddings.

We answer \textbf{RQ-6} by observing Figure \ref{Figure: Training and Inference Time comparison}. It can be observed that the NHP and THP are $10\times$ and $4\times$ slower compared to our TESET model. Even the Hierarchial approach is $1.3\times$ slower. We additionally present asymptotic computation and time complexity analysis in Table \ref{Table: Operations Complexity}.

From Figure \ref{Figure: Features Ablation}, we can compare the training plots for the following models (i) simple transformer, (ii) transformer with distributional heads, (iii) Bayesian transformer with distributional heads. The considerable advantage of the model (iii) is clearly visible from the plots, thus answering \textbf{RQ-7}.

Finally, from Figure \ref{Figure: IntensityAblation}, we can see that the predicted intensities of various correlated diseases are shown to be high in the future. The peak of the curves coincides with the next hospitalization date in the dataset. Thus, not only is \textbf{RQ-8} answered by meaningful predictions, but also the hospitalization can be prevented with precautionary checkup before the predicted date of hospitalization.


\section{Limitations and Future Works}

Our method is limited to representing the relationship among items (such as diseases and treatments) as distances amongst embeddings in the representation space. Consequently, our method can only capture pairwise relationships between items but not more complex relationships such as transitive or hierarchical.
One natural extension of our work would be to infuse external knowledge from a knowledge graph. From the entities and their relationships, it might be possible to capture more complex relationships among items and additional information such as their attributes and side effects.

Another limitation of our method is that it only predicts the items in the set themselves, and not how the set should be used in a decision-making context. For example, if we are predicting diseases and treatments, our method cannot tell us which treatment is best for a particular patient.
Thus, another ambitious future direction for our work would be to extend our work for decision making, for instance by learning decision-making strategies that tell us which treatment to give to a patient at each time step, based on the patient's current state and the history of treatments that they have received. This is often called the dynamic treatment regime, and extending our work in this domain would make it more useful in real-world applications.


\section{Conclusion}

In this paper, we propose a method for modeling the temporal event set distribution. We additionally learn self-supervised contextual event embeddings and incorporate temporal and domain-specific features into the framework to generate better representations. We also provide a Transformer based approach along with SpatioTemporal Encodings to model the same. We empirically demonstrate the validity of our methods along with the necessity of the various components of our proposed methods through appropriate experiments.


\section*{Acknowledgement}
The authors would like to thank the SERB, Department of Science and Technology, Government of India, for the generous funding towards this work through the IMPRINT Project: IMP/2019/000383.

\newpage

%

\bibliography{acml23}

\newpage

\appendix

\begin{center}
    \Large
    \textbf{Deep Representation Learning for Prediction of Temporal Event Sets}\\
    \textbf{in the Continuous Time Domain -- Appendix}
\end{center}

\section{Dataset statistics}
We provide the dataset statistics in Table \ref{Appendix: Dataset Statistics} corresponding to the datasets mentioned in Section \ref{Subsection: Datasets}, which were used in our experiments.
\begin{table}[ht]
    \centering
    \small
    \begin{tabular}{lccc}
    \toprule
     & Synthea & Instacart & MIMIC-III \\
    \midrule
    Total \# data-points & 55299 & 110035 & 1865 \\
    Average seq length & 7.25 & 15.65 & 4.06 \\
    Average set length & 1.19 & 7.23  & 2.79 \\
    \# i/p event types & 211 & 135 & 211 \\
    \# target event types & 124 & 135 & 124 \\
    Dataset type & Synthetic & Real & Real \\
    \bottomrule
    \end{tabular}
    \caption{Dataset statistics.}
    \label{Appendix: Dataset Statistics}
\end{table}

\section{Hyperparameters}
We fix the embedding dimensions, $d_{emb} = 100$ throughout all our experiments, i.e., the item embedding dimension, the output dimension of the item-set embedding generator model, and the hidden dimensions of the transformers are all fixed to be $100$. We additionally fix the number of transformer encoder layers to $2$ and the number of attention-heads to $4$.

\begin{table}[!ht]
    \centering
    \begin{tabular}{lc}
        \toprule
        \textit{Contextual Embedding Encoder} & \\
        \midrule
        Hidden dimension & 100 \\
        Learning Rate & 0.0005 \\
        Dropout & 0.1 \\
        Num Layers & 1 \\
        Batch Size & 128 \\
        \toprule
        \textit{Single-step Model} & \\
        \midrule
        Hidden dimension & 100 \\
        Dense Layer dimension & 256 \\
        Learning Rate & 0.003 \\
        Dropout & 0.1 \\
        Num Encoder Layers & 2 \\
        Batch Size & 512 \\
        Max Seq Length & 500 \\
        Dice $\epsilon$ & 0.1 \\
        Loss $\lambda_1, \lambda_2, \lambda_3$ & $0.85,1,0.2$ \\
        \bottomrule
    \end{tabular}
    \caption{Hyper-parameter table}
\end{table}

\section{Training Details}
For the item embedding generator that we used in Section \ref{SubSection: Self-supervised Pre-training}, we use s single layer of densely connected (feed-forward) neural network as our auxiliary encoder $\mathcal{A}_E$ without any activation at the output (embedding) layer. We use transformers as sequence encoders in the Single-step training approach (Section \ref{Section: Proposed Approaches}). We use transformers with $2$ encoder layers as our sequential encoder while reporting the results. These are not the best possible results, rather, we tried to keep the model's capacity/expressive-power/architecture similar to baselines for a fair comparison.

However, we have experimented with LSTMs as well. It should be noted that due to the bi-directional embeddings in both the Bi-LSTMs and Transformers, a specialized dataset preprocessing is required, which can be skipped if using simple LSTMs. This reduces the training time in LSTMs. However, the inference time remains asymptotically the same.

We use an $80-20$ train-test split in our datasets, and within the training split, we further use $10\%$ of the data for validation. We use Nvidia A100 GPUs to run our experiments.

\section{Notations}

\begin{table}[h]
    \centering
    \small
    \caption{Description of the notations used in the main paper}
    \begin{tabular}{cl}
        \toprule
        \textbf{Notation} & \textbf{Description}  \\
        \midrule
        $\mathcal{S}$ & It defines the input sequence of event sets in continuous time domain \\
        $\mathbf{s}_k$ & It denotes the event set in the input sequence $\mathcal{S}$ \\
        $\mathcal{I}$ & It defines the set of all possible events \\
        $\mathbf{f}_k$ & It denotes the set of features associated with $\mathbf{s}_k$ in the input sequence $\mathcal{S}$ \\
        $\mathbf{t}_k$ & It denotes the timestamp associated with $\mathbf{s}_k$ in the input sequence $\mathcal{S}$ \\
        $\mathcal{T}$ & It defines the set of target events $\mathcal{T} \subset \mathcal{I}$ \\
        $\mathcal{M}$ & It denotes the model being trained for a given task \\
        $\mathcal{A}_E$ & It denotes the auxiliary encoder model \\
        $\mathbf{v}_{emb}$ & It denotes the embedding for an event $\mathbf{i}$ from the auxiliary encoder model $\mathcal{A}_E$ \\
        $\mathbf{d}_{emb}$ & It denotes the dimension of the event embedding from the auxiliary encoder model $\mathcal{A}_E$ \\
        $\mathcal{L}_{aux}$ & It defines the auxiliary contextual loss objective \\
        $\mathcal{H}_k$ & It denotes all the previous set of events along with their corresponding timestamps \\
         & and features until $\mathbf{s}_k$ \\
        $\hat{[ \cdot ]}$ & A hat over any symbol indicates that it is the model's prediction \\
        $\mathbf{e}_{k+1}$ & It denotes the target event set corresponding to the input history $\mathcal{H}_k$ \\
        $\mu_{\hat{\mathbf{e}}_{k+1}}^j$ & It defines the Gaussian distributional parameter - mean of $j^{\text{th}}$ mixture for the target\\
         & event set $\mathbf{e}_{k+1}$\\
        $\sigma_{\hat{\mathbf{e}}_{k+1}}^j$ & It defines the Gaussian distributional parameter - standard deviation of \\
         & $j^{\text{th}}$ mixture for the target event set $\mathbf{e}_{k+1}$\\
        $\alpha_{\hat{\mathbf{e}}_{k+1}}^j$ & It defines the mixture coefficient of $j^{\text{th}}$ mixture for the target event set $\mathbf{e}_{k+1}$ \\
        $\mu_{\hat{\mathbf{t}}_{k+1}}^j$ & It defines the Gaussian distributional parameter - mean of $j^{\text{th}}$ mixture for the target\\
         & time $\mathbf{t}_{k+1}$\\
        $\sigma_{\hat{\mathbf{t}}_{k+1}}^j$ & It defines the Gaussian distributional parameter - standard deviation of \\
         & $j^{\text{th}}$ mixture for the target time $\mathbf{t}_{k+1}$\\     
        $\alpha_{\hat{\mathbf{t}}_{k+1}}^j$ & It defines the mixture coefficient of $j^{\text{th}}$ mixture for the target time $\mathbf{t}_{k+1}$\\
        \bottomrule
    \end{tabular}
    \label{Table: Notations}
\end{table}

\section{Additional Tables}

\begin{table}[h]
    \centering
    \caption{\small \textbf{Additional metrics for Temporal Event-set Modeling Results.} We compare our approaches to baselines, and in addition to Table 1 in the main paper, report the F-scores and RMSEs.}
    \label{Table: Temporal Event Modeling Results2}
    \begin{tabular}{lcccc}
        \toprule
        \multicolumn{1}{c}{\multirow{2}{*}{Training method}} & \multicolumn{2}{c}{Synthea} & \multicolumn{2}{c}{Instacart} \\ \cline{2-5} 
        \multicolumn{1}{c}{} & \begin{tabular}[c]{@{}c@{}}Event-set\\pred (F-score)\end{tabular} & \begin{tabular}[c]{@{}c@{}}Time pred\\ (RMSE)\end{tabular} & \begin{tabular}[c]{@{}c@{}}Event-set\\pred (F-score)\end{tabular} & \begin{tabular}[c]{@{}c@{}}Time pred\\ (RMSE)\end{tabular} \\
        \midrule
        \textit{Baselines:} &  &  &  &  \\
        Neural Hawkes Process & 0.02 & 6.68 & 0.27 & 0.18 \\
        Transformer Hawkes Process & 0.08 & 5.85 & 0.34 & 0.17 \\
        Hierarchical Model & 0.10 & 6.10 & 0.35 & 0.12 \\
        \midrule
        \textit{Ours:} &  &  &  &  \\
        TESET & 0.32 & 4.73 & 0.45 & 0.06 \\
        TESET + Contextual Embeddings & \textbf{0.47} & \textbf{4.28} & \textbf{0.64} & \textbf{0.04} \\
        \bottomrule
    \end{tabular}
\end{table}

\begin{table}[h]
    \centering
    \caption{\small \textbf{Additional Fine-tuning results.} In addition to Table 2 in the manuscript, we report additional metrics: F-score and RMSE, and compare fine-tuning vs training from scratch results for our method vs the baselines.}
    \label{Table: Fine-tuning2}
    \begin{tabular}{clcccc}
        \toprule
        \multirow{2}{*}{FT?} & \multicolumn{1}{c}{\multirow{2}{*}{Training method}} & \multicolumn{2}{c}{Synthea} & \multicolumn{2}{c}{Instacart} \\ \cline{3-6} 
        & \multicolumn{1}{c}{} & \begin{tabular}[c]{@{}c@{}}Event-set\\ given time\\ (F-score)\end{tabular} & \begin{tabular}[c]{@{}c@{}}Time\\ given event\\ (RMSE)\end{tabular} & \begin{tabular}[c]{@{}c@{}}Event-set\\ given time\\ (F-score)\end{tabular} & \begin{tabular}[c]{@{}c@{}}Time\\ given event\\ (RMSE)\end{tabular} \\
        \midrule
        \multirow{4}{*}{\shortstack[c]{Trained \\from \\scratch}} & Neural Hawkes Process & 0.25 & 8.66 & 0.48 & 4.39 \\
        & Transformer Hawkes Process & 0.27 & 7.08 & 0.45 & 3.88 \\
        & Hierarchical Model & 0.24 & 7.81 & 0.44 & 4.50 \\
        & TESET (Ours) & \textit{0.32} & \textit{6.97} & \textit{0.52} & \textit{3.12} \\
        \midrule
        \multirow{4}{*}{\begin{tabular}[c]{@{}c@{}}Fine-\\tuned\end{tabular}} & Neural Hawkes Process & 0.09 & 9.40 & 0.41 & 5.03 \\
        & Transformer Hawkes Process & 0.18 & 7.55 & 0.45 & 4.75 \\
        & Hierarchical Model & 0.16 & 8.13 & 0.48 & 4.99 \\
        & TESET (Ours) & \textit{\textbf{0.44}} & \textit{\textbf{6.27}} & \textit{\textbf{0.60}} & \textit{\textbf{2.25}} \\
        \bottomrule
    \end{tabular}
\end{table}

\newpage

\section{Additional Figures}
\begin{figure}[h]
    \centering
    \includegraphics[width=\textwidth]{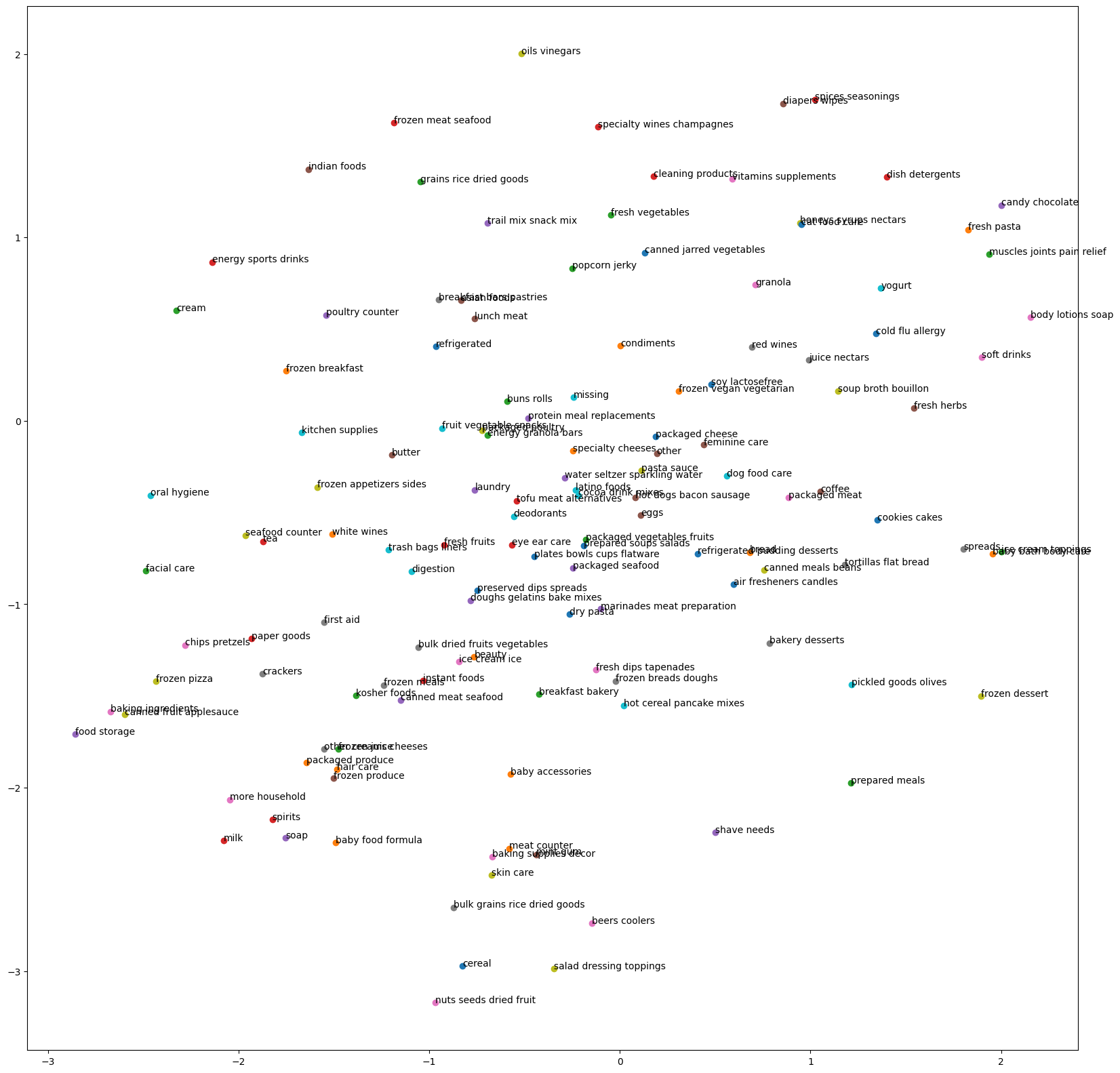}
    \caption{\small 2D t-SNE of the embedding space after the first step of training (learning the contextual representations) for the Instacart Dataset. It can be observed that the clusters formed in the embedding space are valid. For instance, frozen meals and instant noodles form a cluster, soy, and vegan items are in the same cluster, and bakery and dough are also in close proximity to each other.}
    \label{fig:tsne}
\end{figure}

\begin{figure}[h]
    \centering
    \includegraphics[width=1.2\textwidth]{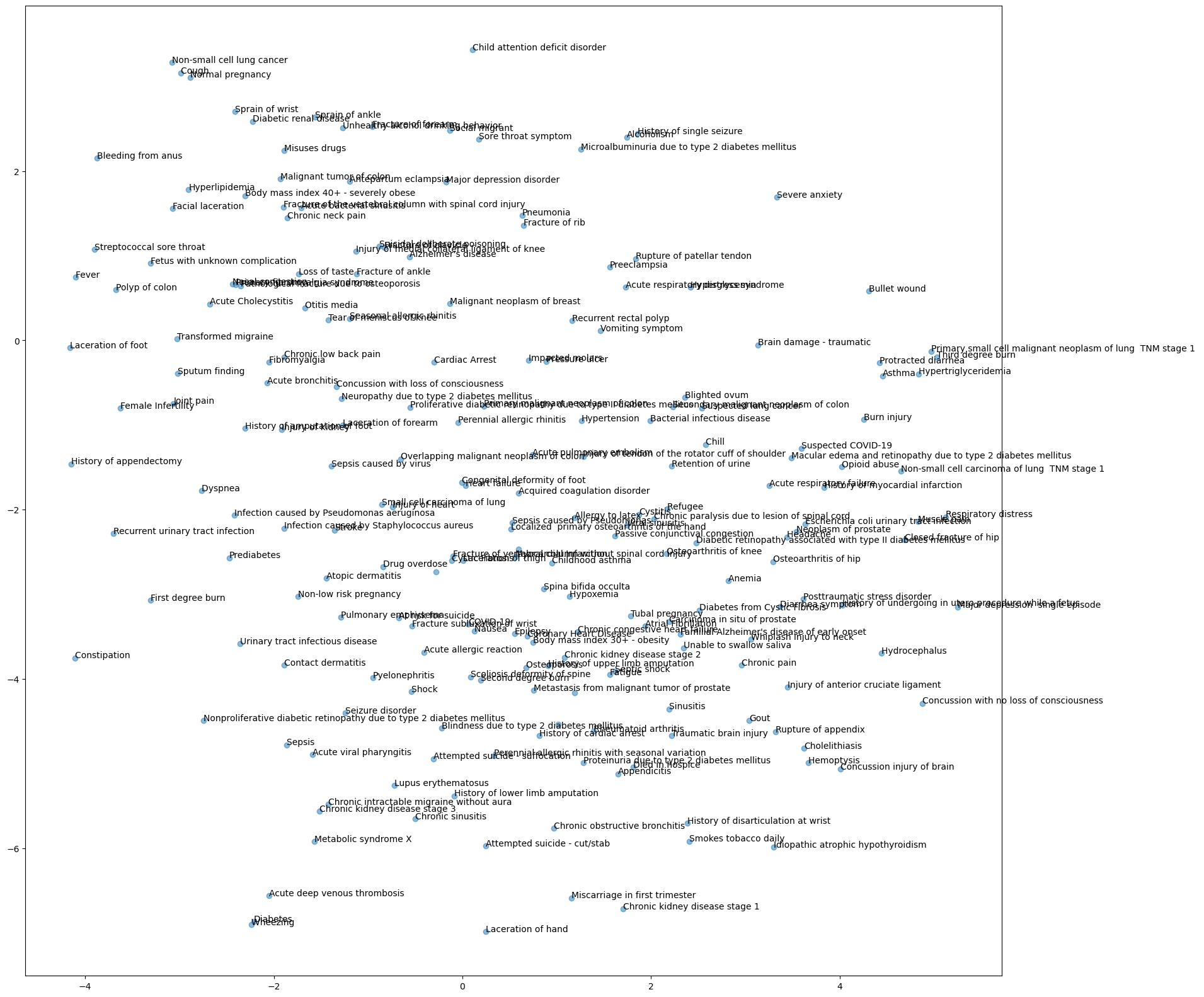}
    \caption{\small 2D t-SNE (full) of the embedding space after the first step of training (learning the contextual representations) for the Synthea Dataset.}
    \label{fig:tsne_disease}
\end{figure}

\end{document}